%% file: knowledgeable-snippet.tex
\documentclass[11pt,a4paper]{article} 
\usepackage{amssymb}
\usepackage{graphicx}
\usepackage{bm}
\usepackage{colortbl}
\usepackage{subfigure}
\usepackage{amsmath}
\usepackage{amsfonts}
\usepackage{pifont}
\usepackage{url}
\usepackage{multirow}
\usepackage{hyperref}  
\definecolor{mygray}{gray}{.9}
\newcommand{\definen}[3]{\textbf{#1}~\textbf{#2}~(\textbf{#3}).}

\usepackage[top=1in,bottom=1in,left=1in,right=1in]{geometry}

\date{}

\begin{document}

\title{Hierarchical Neural Network for \\ Extracting Knowledgeable Snippets and Documents}

\author{Ganbin Zhou$^{1,2,3}$, Rongyu Cao$^{1,2}$, Xiang Ao$^{1,2}$, Ping Luo$^{1,2}$, \\ Fen Lin$^{3}$, Leyu Lin$^{3}$, Qing He$^{1,2}$ \\
	$^{1}$ Key Lab of Intelligent Information Processing of Chinese Academy of Sciences (CAS), \\Institute of Computing Technology, CAS, Beijing 100190, China. \\
	$^{2}$University of Chinese Academy of Sciences, Beijing 100049, China. \\
	$^{3}$ Search Product Center, WeChat Search Application Department, Tencent, China.\\
}

\pdfinfo{
	/Title (Mechanism-Aware Neural Machine for Dialogue Response Generation)
	/Author (Ping Luo, Ganbin Zhou, Rongyu Cao, Fen Lin, Bo Chen, Qing He) }

\maketitle

\input{abstract/abs_v1_zgb-ao}
\input{introduction/v3-xiang}
\input{relatework/v3-xiang}
\input{content/definition_zgb_xiang_v2}
\input{content/ssnn-v4_zgb-ao}

\input{content/svm-cry-xiang_v2}

\input{content/exp_zgb_xiang_v2}

\input{conclusions/v1_zgb}

\bibliographystyle{plain}
\bibliography{docs}

\end{document}

%% file: abstract/abs_v1_zgb-ao.tex
\begin{abstract}
In this study, we focus on extracting knowledgeable snippets and annotating knowledgeable documents from Web corpus, consisting of the documents from social media and We-media. Informally, knowledgeable snippets refer to the text describing concepts, properties of entities, or relations among entities, while knowledgeable documents are the ones with enough knowledgeable snippets. These knowledgeable snippets and documents could be helpful in multiple applications, such as knowledge base construction and knowledge-oriented service. Previous studies extracted the knowledgeable snippets using the pattern-based method. Here, we propose the semantic-based method for this task. Specifically, a CNN based model is developed to extract knowledgeable snippets and annotate knowledgeable documents simultaneously. Additionally, a ``low-level sharing, high-level splitting'' structure of CNN is designed to handle the documents from different content domains. Compared with building multiple domain-specific CNNs, this joint model not only critically saves the training time, but also improves the prediction accuracy visibly. The superiority of the proposed method is demonstrated in a real dataset from Wechat public platform.

\end{abstract}

%% file: introduction/v3-xiang.tex
\section{Introduction}

\begin{figure}[htbp]
	\centering
	\subfigure[The document introduces the turning skills of driving. 
    ]{
		\fbox{\includegraphics[width=2.9in]{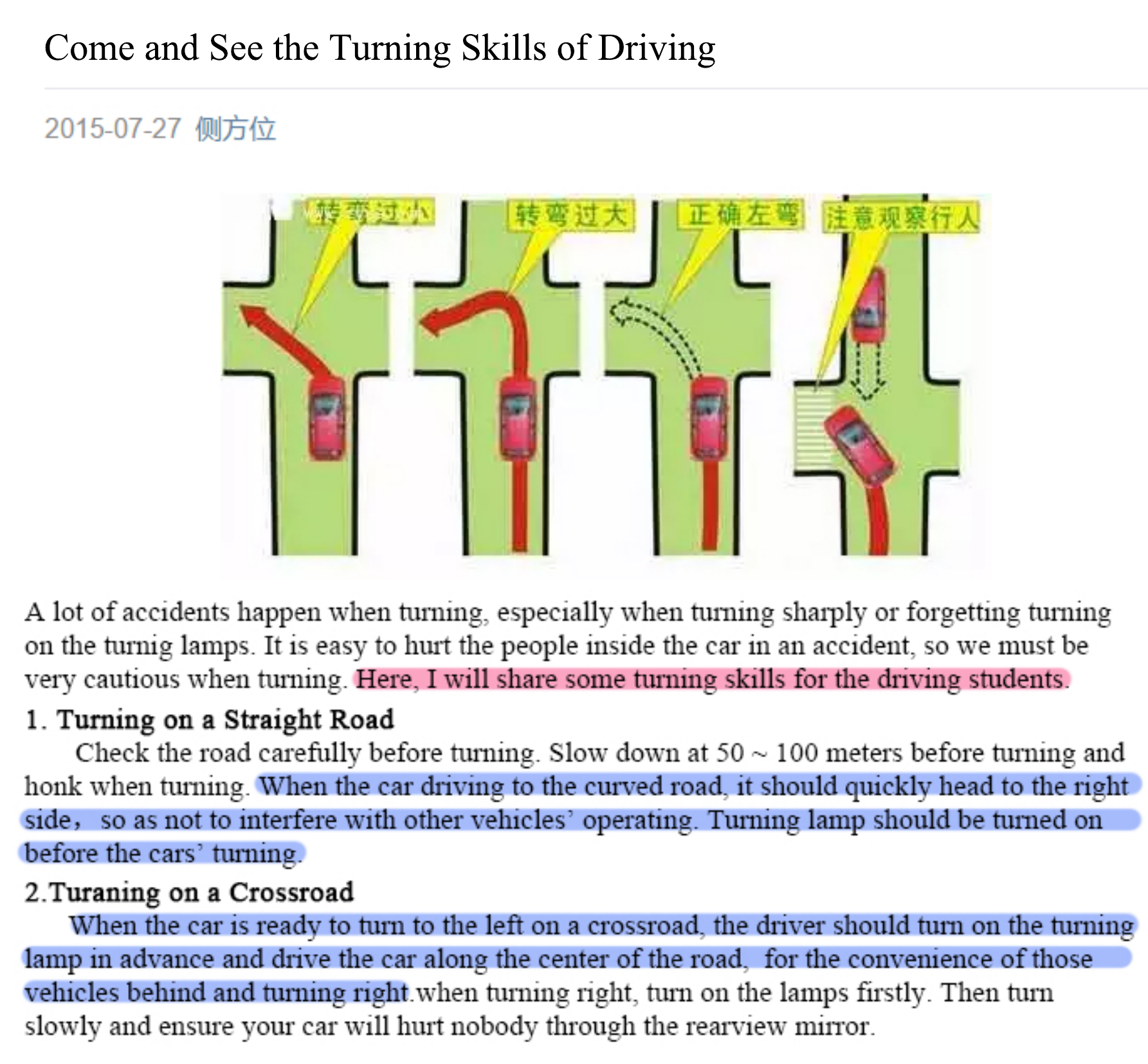}}
		\label{fig:screenshots:b}}~~~~
	\subfigure[The document introduces the 25 pieces of tips for purchasing real estates. 
        ]{
		\fbox{\includegraphics[width=2.9in]{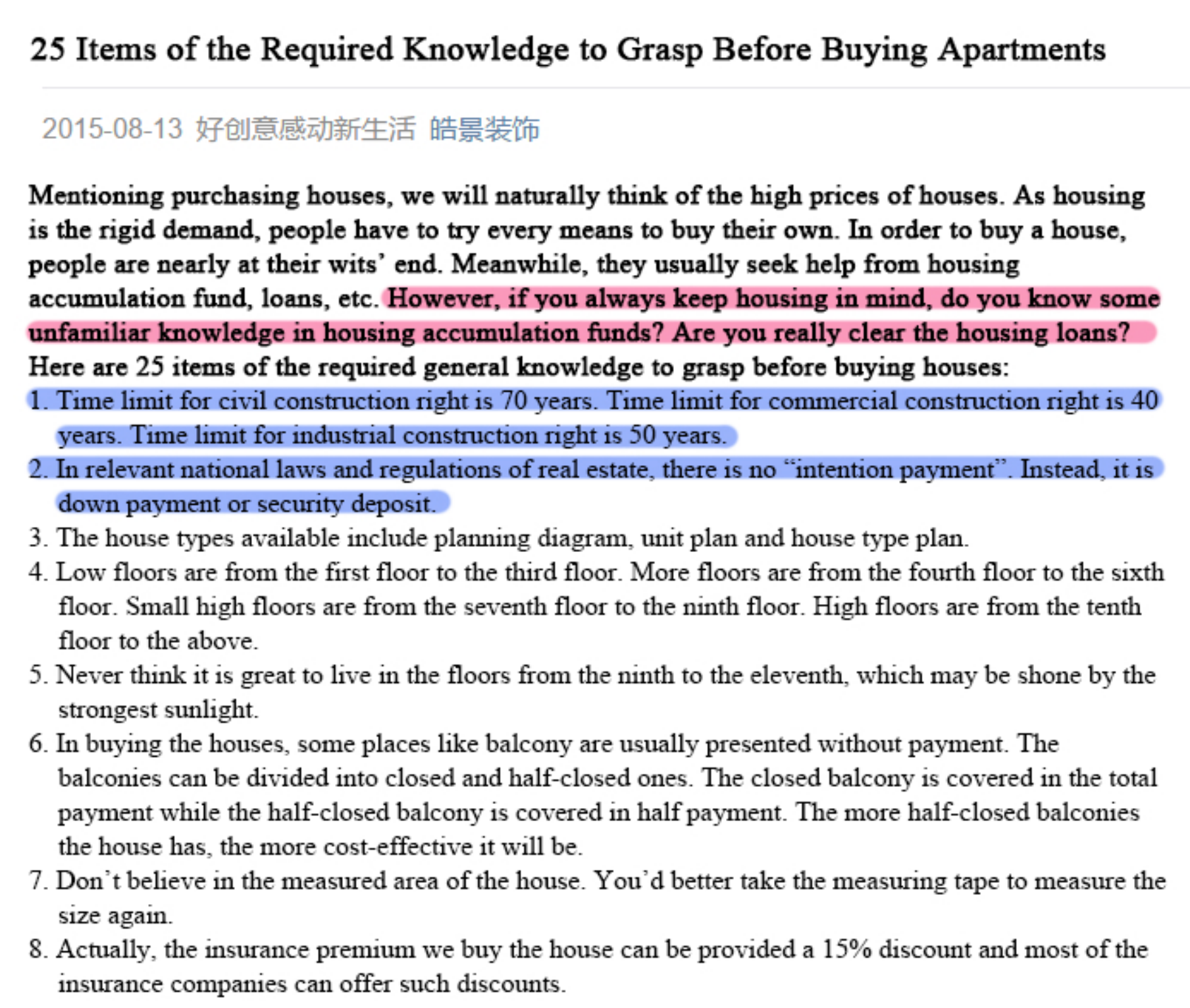}}
		\label{fig:screenshots:c}}
	\caption{Examples of knowledgeable documents. The blue and red sentences are knowledgeable and unknowledgeable snippets respectively.
}   
	\label{fig:screenshots} 
\end{figure}

Nowadays millions of articles are published in social media everyday. Some of them attract millions of clicks, forwardings, and favorites because of their high-quality contents, which usually provide various knowledge and experiences from different domains.
For example, Figure~\ref{fig:screenshots} exhibits two screenshots of such articles from Wechat public platform. The first one, shown in Figure~\ref{fig:screenshots:b}, is a popular article introducing the turning skills of driving, while the other article in Figure~\ref{fig:screenshots:c}
summaries the $25$ tips for purchasing real estates. These two documents, due to their usefulness, receive more than $100,000$ views in several days, respectively.

By being able to recognize ``knowledgeable'' documents as well as their ``knowledgeable'' snippets from large web corpus helps in wide potential applications. We describe two of such applications. First, knowledgeable articles and snippets can be used as a data source for knowledge base construction.
Currently, majority of popular knowledge bases, like YAGO~\cite{suchanek2007yago}, DBpedia~\cite{auer2007dbpedia} and etc., extracted knowledge based on Wikipedia, WordNet, GeoNames and so on. Compared with the data scale of social media, knowledge in these structured or semi-structured resources summarized by human beings might be limited and inflexible. Another recent knowledge base, Probase, with $2.7$ million concepts was automatically harnessed from the so-far largest corpus consisting of $326$ million knowledgeable sentences extracted from $1.68$ billion web pages~\cite{Wu2012Probase}. However, these sentences are extracted only by the Hearst patterns~\cite{Hearst1992}. For extracting more knowledgeable snippets to construct more comprehensive knowledge base, semantic-based methods are needed to complement the previous pattern-based ones.


The second potential application is knowledge-oriented services, e.g. knowledge retrieval and question-answering. In particular, these extracted knowledgeable documents and snippets can be used directly as the answers to the questions raised by users when they need some help a certain issue. For example, if a user wants to know some knowledge on purchasing real estates, then the article as shown in Fig~\ref{fig:screenshots:c} can be retrieved by these knowledge-oriented systems.

Motivated by these potential applications, we investigate the problem of annotating knowledgeable documents and extracting their knowledgeable snippets from large-scale web corpus. Informally, a knowledgeable document is a document containing multiple knowledgeable snippets, which describe concepts, properties of entities, or the relations among entities. In this study, we address this task by analyzing the semantics of document text based on the convolutional neural network~(CNN) model. CNN has been applied in understanding images and natural language text in recent years~\cite{zeiler2010deconvolutional,zeiler2011adaptive}, and achieved great success in multiple applications, such as image recognition and captioning~\cite{Krizhevsky,Vinyals2014,Xu2015}, text sentiment analysis~\cite{Kalchbrenner2014,Denil2015}, non-photorealistic rendering~\cite{Gatys2015}, etc..

Specifically, we propose SSNN, a joint CNN-based model, to understand the abstract concept of documents in different domains collaboratively and judge whether a document is knowledgeable or not. In more detail, the network structure of SSNN is ``low-level Sharing, high-level Splitting'', in which the low-level layers are shared for different domains while the high-level layers beyond the CNN are trained separately to perceive the differences of different domains. It is an end-to-end solution for document annotation without the time-consuming feature engineering work. In addition, we carefully develop the manual features for this task and train a SVM classifier. 
We conduct extensive experiments on the real documents from three content domains in Wechat public platform to demonstrate the superiority of the proposed SSNN.

The contributions of this study are summarized as follows:
\begin{list}{$\bullet$}
{ \setlength{\leftmargin}{2em}}
    \item We formulate the problem of knowledgeable documents and snippets extraction. 

    \item We propose a ``low-level \textbf{S}haring, high-level \textbf{S}plitting'' \textbf{N}eural \textbf{N}etwork (SSNN for short) to identify knowledgeable documents and annotating knowledgeable snippets simultaneously.
    \item We carefully design the manual features for the knowledgeable documents and snippets extraction task and train a SVM classifier.

    \item We verify the performance of the proposed models on a real dataset from Wechat public platform. The results show that the proposed SSNN is a promising solution towards on knowledgeable document and snippets extraction.
\end{list}

The remainder of this paper is organized as follows.
Section~\ref{sec:related} surveys the related work.
Section~\ref{sec:definition} presents the problem formulation.
Section~\ref{sec:cnn} details the proposed SSNN model.
Section~\ref{sec:svm} details the SVM based method with the manual features. 
Section~\ref{sec:experiment} demonstrates the experimental results.
Section~\ref{sec:conclusion} concludes the paper.

%% file: relatework/v3-xiang.tex
\section{Related Work} \label{sec:related}



\textbf{Genre Identification}.
Our problem is related to the genre classification problem.
In genre classification, documents are mainly divided into several types, e.g. narrative, exposition and so on. 
Though exposition document generally provides the background information, concepts or properties of things, the proposed knowledgeable documents also discuss relations, cause and influence between entities, which include more broad definitions than the exposition documents.

The previous work about genre identification mainly contribute in two aspects: (1) Find more effective features to represent the document; (2) Propose more advanced machine learning algorithms for classification.

Kessler et al.~\cite{kessler1997automatic} first used the term ``genre'' to represent any widely recognized class of texts defined by some common communicative purposes or other functional traits.
Finn et al.~\cite{finn2006learning} proposed that the genre classification is orthogonal to the topic classification. That is to say the documents had same topic can have different genre and documents in the same genre can also have different topics.
Feldman et al.~\cite{feldman2009part} put forward a method for part of speech feature extraction and obtain a significant improvement for genre identification.
Denil et al.~\cite{Denil2015} proposed an advanced machine learning algorithm to represent the text into a lower dimension. Such a representation can characterize the inter-class separability and intra-class compactness by the special designed intrinsic graph and penalty graph. 


\textbf{NLP via Neural Network}. 
Knowledgeable document extraction is related to NLP classification.
For such task, Tai et al.~\cite{DBLP:journals/corr/TaiSM15} proposed a model that predict the semantic relatedness of two sentences and sentiment classification based on LSTM model.  Kim et al.~\cite{DBLP:journals/corr/Kim14f} proposed a CNN model based on hyperparameter tuning and static vectors for sentence classification.
However, these models focus on the sentence level classification, while our work focuses on the classification in document level.
Moreover, knowledgeable snippet extraction is related to the task of identifying task-specific sentences. For such research field, Denil et al.~\cite{Denil2015} proposed a hierarchical convolutional model which can be used for documents classification and topic-relevant sentences.
Tu et al.~\cite{Tu2012Identifying} proposed a method which automatically identify high impacting sub-structures. However, both of these methods are designed for corpus consisting of documents from a same topic. While in our work, documents are come from large-scale corpus of social media with many topics.


%
%


%% file: content/definition_zgb_xiang_v2.tex
\section{Problem Formulation}\label{sec:definition}
In this section, we first introduce the informal definitions of knowledge snippet and knowledgeable document, and then formulate the learning problem.


\definen{Definition}{1}{Knowledgeable Snippet} A snippet of a document is called \emph{knowledgeable} if it conforms to one of the following descriptions.

\begin{enumerate}
	\item \textbf{The \emph{definition} of a concept or an entity.}

For example, ``\emph{A company is an association or collection of individuals, whether natural persons, legal persons, or a mixture of both}.'' and ``\emph{Google Inc. is an American multinational technology company specializing in Internet-related services and products.}'' are both knowledgeable snippets since they define the concept of ``company'' and the entity of ``Google Inc.''.
Note that concept also refers to abstract entity. Hence for convenience and clarity, we use entity as a unified notion hereafter.

\item \textbf{The \emph{property} of an entity. }

For example, ``\emph{Company members share a common purpose and unite in order to focus their various talent.}'' is knowledgeable. Here ``members'' is a property of the entity ``company''.

\item \textbf{The \emph{relation} between two entities. }

For example, ``\emph{Because companies are legal persons, they also may associate and register themselves as a corporate group.}'' is knowledgeable because the sentence discusses the relation between ``company'' and ``corporate group''. They are both entities.

\item \textbf{The \emph{cause} of a relation between entities.}

The previous example is also proper for this description.
Specifically, ``companies are legal persons'' is the cause of the relation between entity ``company'' and ``corporate group''.

\item \textbf{The \emph{influence} of the relation between entities.}

    For example, ``\emph{The members guarantee the payment of certain amounts if the company goes into insolvent liquidation}'' is a knowledgeable sentence, and ``the members guarantee the payment of certain amounts'' is the influence of ``company goes into insolvent liquidation''.

\end{enumerate}

Generally speaking, knowledgeable is highly related to the definition of entities, properties of entities and relations between entities. The examples used in Definition~1 are selected from Wikipedia.


\definen{Definition}{2}{Knowledgeable Document} A document is called \emph{knowledgeable} if more than half of snippets in it are knowledgeable.


These informal definitions on knowledgeable snippets and documents give us some intuitive understandings on these concepts. They are also used as a guideline for the volenteers to label the training data set for this task.

Based on the definitions above, we formulate our learning problem of extracting knowledgeable documents and snippets as follows:



\textbf{Learning Problem:} Given $Q$ domains of documents $\mathcal{D}^{1}$, $\mathcal{D}^{2}$, $\dots$, $\mathcal{D}^{Q}$, the labeled document set from the $q$-th domain is denoted as $\mathcal{D}_{l}^{q}$$=$$\{(x_{l}^{q}, y_{l}^{q})$$| y_{l}^{q}$ $\text{ is supervising label of }$$ x_{l}^{q}\}$, where $x_{l}^{q}$ denotes a document in $q$-th domain with supervising label, and $y^{q}_{l}$ is a binary label ($0$ or $1$) indicating knowledgeable or unknowledgeable. The unlabeled document set from the $q$-th domain is denoted as $\mathcal{D}_{u}^{q}=\{x_{u}^{q}\}$ where $x_{u}^{q}$ denotes a document in $q$-th domain without supervising label. This task is to predict the label for a unlabeled document, and meanwhile annotate its knowledgeable snippets.

%% file: content/ssnn-v4_zgb-ao.tex
\section{The SSNN Model}\label{sec:cnn}



In this section, we introduce the proposed SSNN model that identifies knowledgeable documents from corpus and annotate knowledgeable snippets of them.






Generally speaking, SSNN is designed to improve the generalization ability and decrease the time consuming of training.
In SSNN, the low-level layers are shared while the high-level structures are split for multiple domains. This idea is motivated by the structure of human brain, and we take the vision system of human as an example. When an image is shot into the human's eyes, the primary visual cortex firstly transforms elementary information processed by neural cells into abstract features. At this stage, information is always sharing. Then, the information begins to be split and is fed to different types of advanced cortex and let people recognize the picture from different angles~\cite{zeki1993vision}.

Similar to the human vision, in the proposed SSNN, CNN layers are adopted to handle low-level features, namely words and sentences.
The CNN layers are shared among domains, corresponding to the neural system connecting eyes and primary visual cortexes. The output of low-level layers is fed to split domain-specific high-level layers (e.g. softmax layers), which is corresponding to the advanced cortex as the vision example.
\subsection{The Structures of SSNN}


To predict whether a given document is knowledgeable or not, SSNN will: (1) transform the document into a document embedding; (2) feed the document embedding
to a softmax layer; (3) use the softmax layer to predict whether a document $x$ is knowledgeable or not.

\begin{figure*}[htbp]
	\centering
	\includegraphics[width=6.2in]{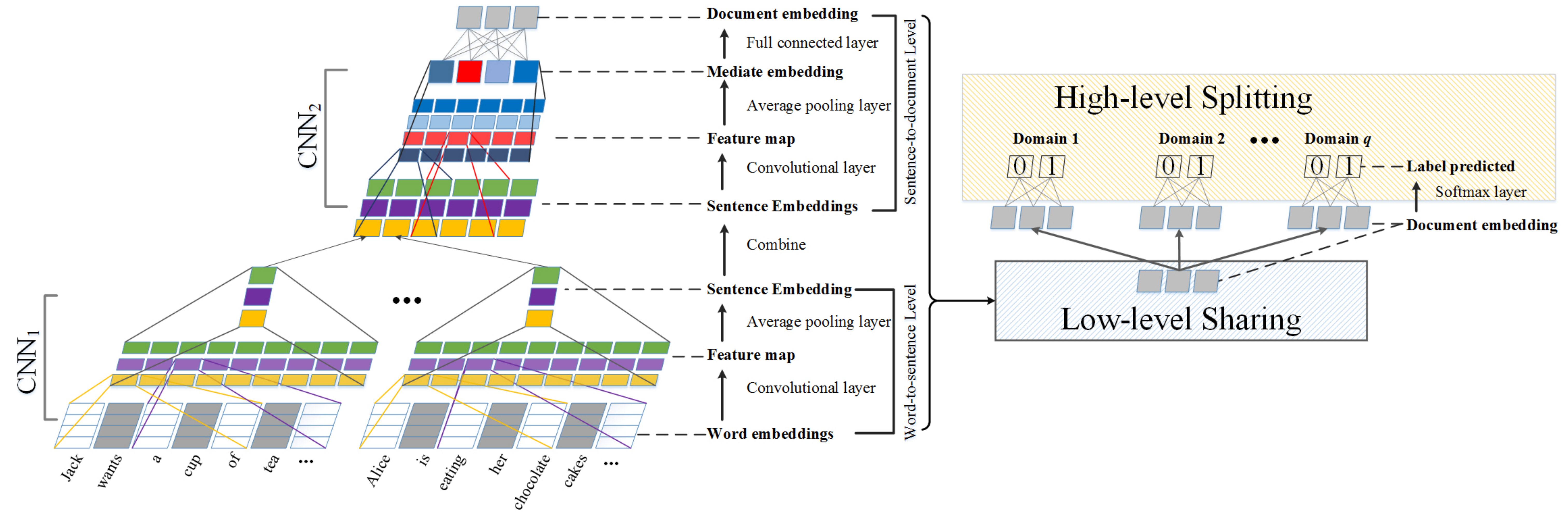}
	\caption{The structure of the proposed model.}
	\label{fig:nnstructure}
\end{figure*}

The structure of SSNN is shown in the Fig.\ref{fig:nnstructure}. The components used for embedding a document into a vector is divided into two levels: word-to-sentence level and sentence-to-document level.

At the word-to-sentence level, we transform words into sentence embeddings. Hence, we first generate word embeddings as the input by applying the word2vec~\cite{Mikolov2013Distributed} and setting dimension to $200$. The vocabulary we used contains $272,582$ Chinese words, and the out-of-vocabulary words are replaced with a special token ``UNK''.
Then, for a sentence of the given document, the embeddings of its words compose a word matrix $\bm{W}=(\bm{w}_1, \bm{w}_2, \cdots, \bm{w}_{|\bm{W}|})$, where $\bm{w}_i(i=1,2,\cdots,|\bm{W}|)$ is the embedding of the $i$-th word. $\bm{W}$  is then transformed into a sentence embedding $\bm{s}$ by CNN$_1$. 
Similarly, at sentence-to-document level, we will transform all sentence embeddings of one document into a document embedding. The embeddings of sentences compose a sentence matrix $\bm{S}=(\bm{s}_1, \bm{s}_2, \cdots, \bm{s}_{|\bm{S}|})$, where $\bm{s}_i(i=1,2,\cdots,|\bm{S}|)$ is the embedding of the $i$-th sentence in a document $x$. $\bm{S}$ will be transformed into a mediate embedding $\bm{d}'$ by CNN$_2$.
The mediate embedding $\bm{d}'$ will be flattened and then fed to a fully connected layer to generate the document embedding $\bm{d}$. After that, $\bm{d}$ is fed to a domain-specific softmax layer to generate prediction. The softmax layer will generate the probabilities of a document $x$ is knowledgeable. Here, if the probability is above $0.5$, $x$ will be predicted as a knowledgeable document. Otherwise, it will be unknowledgeable.

\subsection{Training Details}\label{sec:CNNLayer}




Each convolutional layer of CNN$_{1}$ and CNN$_{2}$ contains a filter bank $\bm{F}\in \mathbb{R}^{a\times q_{\bm{F}} \times n_{\bm{F}}}$, where $a$ is the dimension of input embeddings, $q_{\bm{F}}$ is the filter width and $n_{\bm{F}}$ is the number of convolutional kernels in the filter. Since the dimension of first axis of convolutional kernel and input embedding matrix are the same, every embedding matrix are transformed into vector by convolution operation. We then combine these vectors column by column into a feature map, and feed the feature map to pooling layer.

Note that the lengths of sentences and documents are different. Hence, the column number of feature map generated from sentences and documents are different, and the dimension of the mediate embedding $\bm{d}'$ will vary for different documents. It will be problematic for the fully connected layer which only receiving fixed-size embeddings.
To ensure the dimension of sentence and document embeddings unchanged for different sentences and documents, we average each row of feature map via an average pooling layer which computes the average values along each row of embedding matrix.
When training, SSNN adapts model parameters to minimize a loss function $L(\cdot)$, which is the cross-entropy of the predicted and true labels. In detail, for domain $q (q = 1,2,\cdots,Q)$, let $\theta^q$ denote the parameter set including the word-to-sentence layer, sentence-to-document layer and the softmax layer for domain $q$. Let  $h_{\theta^q}(x_l^q)$ denote the  predicted probability of that $x_l^q$ is knowledgeable. The loss function $L$ for $(x_l^q,y_l^q)$ is defined as follows:
\begin{equation}
\begin{aligned}
L(y_l^q,h_{\theta^q}(x_l^q))=-[y_l^q\log{h_{\theta^q}(x_l^q)}+(1-y_l^q)\log{(1-h_{\theta^q}(x_l^q))}]
\end{aligned}
\end{equation}

Furthermore, the objective function to be minimized is as follows:
\begin{equation}
\min_{\theta^1,\theta^2,\cdots,\theta^Q}~\sum_{q=1}^{Q}~\sum_{(x_{l}^{q}, y_{l}^{q})\in\mathcal{D}_{l}^{q}}
L(y_l^q,h_{\theta^q}(x_l^q))
\end{equation}


To minimize the objective function, we adopt mini-batch gradient descent technique~\cite{NIPS2011_4432} which feeds the model a fixed number of training data in each mini-batch update. This technique makes it possible for memory-limited GPUs to deal with big data set. Here, the data in a mini-batch are only chosen from an individual domain. Given a mini-batch of documents $M=\{(x_l^q,y_l^q)\}$, the parameters are updated as follows:
\begin{equation}
\theta^q_{\mathrm{new}}=\theta^q_{\mathrm{old}}-\frac{\gamma}{|M|}\sum_{(x_{l}^{q}, y_{l}^{q})\in M}
\frac{\partial L(y_l^q,h_{\theta^q_{\mathrm{old}}}(x_l^q))}{\partial \theta^q_{\mathrm{old}}}
\end{equation}
where $\gamma$ is the learning rate, and we set $\gamma=0.1$ . Additionally, since the softmax layers are independent for different domains, only parameters in $\theta^q$ are updated when using a mini-batch $M \subset \mathcal{D}_{l}^{q}$	.


\subsection{Extracting Knowledgeable Sentences}\label{sec:extracSentences}
So far we have discussed how to identify the knowledgeable documents from the corpus. Now we move to extract the knowledgeable snippets based on the proposed SSNN.


Specifically, we accomplish this task by adapting the method of \cite{zeiler2011adaptive}. The basic idea is that a sentence may tend to be knowledgeable if its document becomes less knowledgeable after the sentence is removed.

In particular, for a document $x$, we first predict whether $x$ is knowledgeable or not. We then construct a pseudo-label $\widetilde{y}$ which is the invert of the prediction. For example, if document $x$ is predicted to be knowledgeable , its pseudo-label will be unknowledgeable.

Let $\bm{S}$ denote the sentence matrix of $x$, $\hat{y}=f(\bm{S})$ denote the predicting label output by softmax layer, and  $\bm{I}^{\bm{S}}$ denote an identity matrix whose width is equal to the sentence number of $x$. It can be inferred that $\hat{y}=f(\bm{S})=f(\bm{S}\bm{I}^{\bm{S}})$.
Note that evaluating the influence in the ``knowledgeable'' level of $x$ if the $i$-th sentence of $x$ is removed, is equal to evaluating the influence in $L(\widetilde{y},f(\bm{S}\bm{I}^{\bm{S}}))$ if we transform the $1$ into $0$ at $\bm{I}_{ii}^{\bm{S}}$. We define this influence as the derivative of the loss function $L(\widetilde{y},f(\bm{S}\bm{I}^{\bm{S}}))$ with respect to $\bm{I}_{ii}^{\bm{S}}$, which is as follows:



\begin{equation}\label{eq:formula}
\begin{aligned}
\frac{\partial L(\widetilde{y},f(\bm{S}\bm{I}^{\bm{S}}))}{\partial \bm{I}_{ii}^{\bm{S}}}= \frac{\partial L(\widetilde{y},f(\bm{S}\bm{I}^{\bm{S}}))}{\partial \bm{S}\bm{I}_i^{\bm{S}}} \times \frac{\partial \bm{S}\bm{I}_i^{\bm{S}}}{\partial \bm{I}^{\bm{S}}_{ii}}=\bm{\delta}^{\mathrm{T}}_i\bm{S}_i
\end{aligned}
\end{equation}
where $\bm{S}_i$ is the $i$-th column of the document matrix $\bm{S}$, $\bm{I}^{\bm{s}}_i$ is the $i$-th column of $\bm{I}^{\bm{s}}$, and $\bm{\delta}_i$ is the backpropogation message to $\bm{S}_i$~\cite{Denil2015}.

The derivative in Eq.~\eqref{eq:formula} is defined as the knowledgeable scores for each sentence. If the document $x$ is predicted knowledgeable, the higher score of a sentence indicates the more knowledgeable is the sentence. But if the document $x$ is predicted unknowledgeable, the higher score indicates the sentence is the less knowledgeable~\cite{baehrens2010explain}. Finally, we choose sentences with top-$k$ knowledgeable scores as the knowledgeable snippets for a given document.

\subsection{Analysis on Memory Consumption of SSNN}
In this subsection, we analyze the memory consumption of SSNN and demonstrate that SSNN can save memory consumption compared with traditional CNN models.

Let $n^L$ denote the parameter numbers of the word-to-sentence and the sentence-to-document levels, and let $n^H$ denote the parameter numbers of the softmax layer. Hence, the parameter number of SSNN is $(n^L+Q\cdot  n^H)$ where $Q$ indicates the number of domains. For fair comparison,
we assume that the structure of the domain-specific CNN is the same as one-domain SSNN. Hence, the parameter numbers of a domain-specific CNN is $(n^L+ n^H)$.
To train CNNs for $Q$ domains, the  number of all parameters is $(Q\cdot n^L+Q\cdot n^H)$.
Let $r$ denote the saving ratio of memory consumption of SSNN w.r.t. domain-specific CNNs. $r$ can be derived as follows:
\begin{equation}
r=1-\frac{ n^L+Q\cdot  n^H}{Q\cdot n^L+Q\cdot n^H}
=(1-\frac{1}{Q})\cdot\frac{n^L}{n^L+n^H}
\end{equation}

In real application, since $n^L$ denotes the parameter number of several complex layers of neural network, and $n^H$ only denotes the parameter number of a binary classifier, we obtain $n^L \gg n^H$.
Thus, the saving ratio $r$ can be approximated as $(1-\frac{1}{Q})$. This ratio is impressive, for example, when $Q=3$, SSNN can save $66.67\%$ parameters w.r.t. 3 domain-specific CNNs.
Thus, we argue that the proposed SSNN has benefits on both time and memory consumptions because of its sharing and splitting structure.


%% file: content/svm-cry-xiang_v2.tex

\section{Feature Engineering Solution}\label{sec:svm}

In this section, we introduce another approach to knowledgeable document recognition based on feature engineering.
Specifically, we train a SVM classifier for every domain with manually extracted features and use these classifiers to predict whether the unlabeled documents is knowledgeable or not. Next we will mainly focus on describing the features we adopt in such method.
We divide the features into multiple categories, i.e. POS features, word features and sentence features. Part of speech (POS) features are based on the word POS tagging, word features are related to the meaning of words, and sentence features consider the comprehensive characteristics in sentence level.

\subsection{POS features}

According to Definition~1, knowledge is highly related with entities which are usually noun including general noun~(for abstract concept) or proper noun~(for concrete entity). Besides, when explaining definitions, properties or relations of entities, illustrative text with strong expositions are usually included. Hence noun, verb and verb noun might act as the leading role in knowledgeable sentences.


Review the narrative articles, which are considered unknowledgeable in this study, usually contain news events.
In these documents, characters' names, time words and location words accompanied with a great quantity of verbs might appear frequently because they might describe who did what in when and where.


In addition, some unknowledgeable documents contain lyrical and critical text. Most of them express self views or comments on other things, which are highly subjective. Hence pronoun, interjection, adverb and adjective often appear together in these articles.

\begin{figure}[htbp]
	\centering
	\includegraphics[scale=0.4]{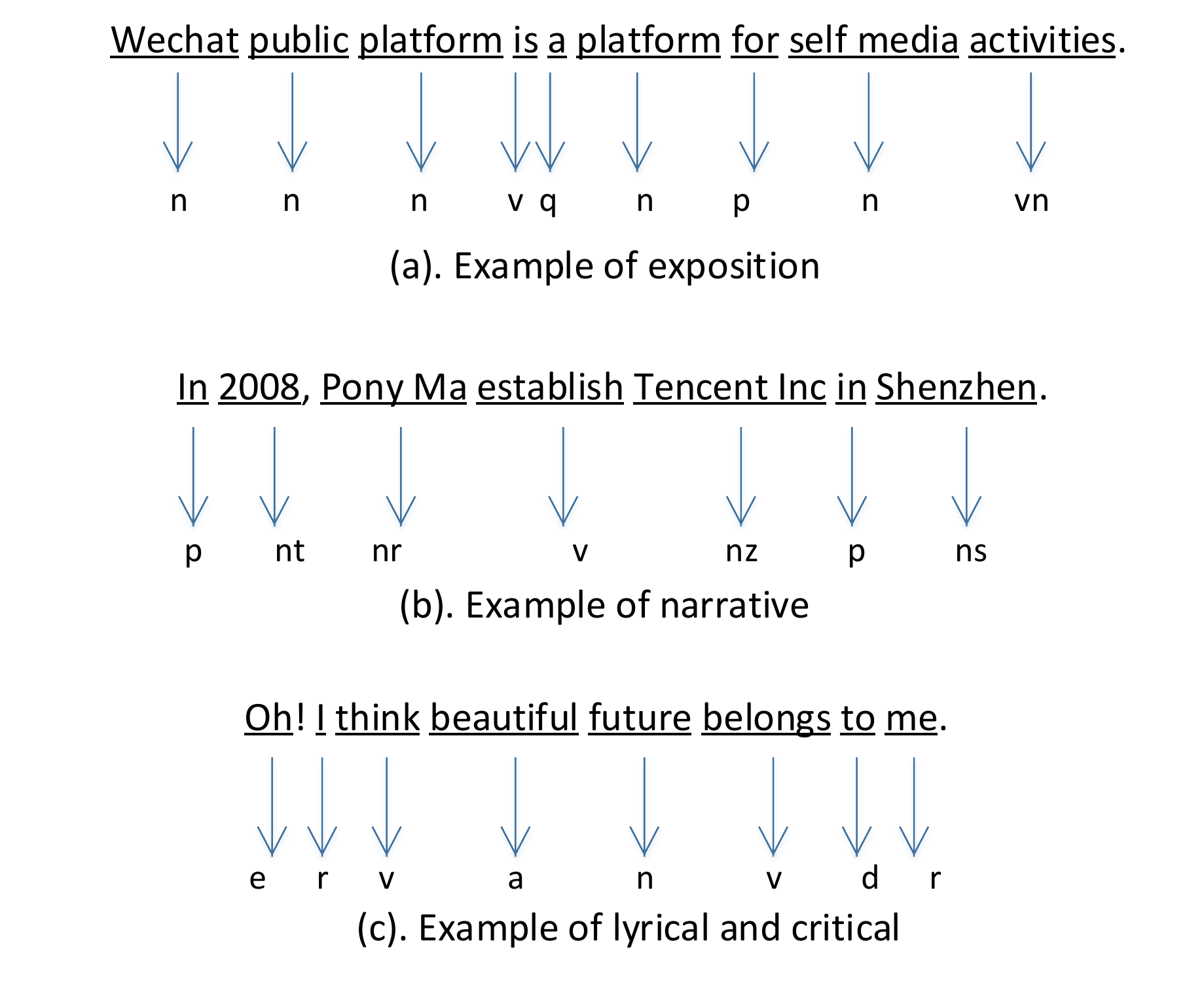}
	\caption{Examples for different POS.}\label{fig:FeatureExamples}
	
\end{figure}

Figure \ref{fig:FeatureExamples} gives some examples we have discussed above. In order to discriminate knowledgeable and unknowledgeable documents, some additional POS features are also needed. The detailed information of adopted POS features is shown as Table~\ref{table:ReasonofUsingPOS}.

\begin{table}[htbp]
	\centering
	\caption{POS pattern in different type of documents.}\label{table:ReasonofUsingPOS}
	\begin{tabular}{l|l|l}
		\hline
		Type & POS & Knowledgeable \\
		\hline
		exposition & general noun, proper noun, verb and verb noun & Yes \\
		\hline
		scientific  & proper noun, quantifier and numeral & Yes \\
		\hline
		narrative & characters' name, time words, location words and verb & No \\
		\hline
		critical & pronoun, interjection, adverb and adjective & No \\
		\hline
		lyrical & idiom, exclamation, abbreviation and onomatopoetic & No \\
		\hline
		advertisement & website url, telephone and email & No \\
		\hline
	\end{tabular}
\end{table}

By analyzing amount of examples, we observe the POS patterns appear with great regularities, so we take the POS histogram statistic features~(PHSF)\cite{feldman2009part} as the POS features. That is we first tag each word in the document by a POS tagger with $k$ POS tags.
We use a sliding window of length $t$ and slide it from the start to the end of a document to get all together $l-t+1$ windows. For every tag $j$, we count the number of tag $j$ in each window and get a sequence $S_{j}$ $=$ $[n^j_1,n^j_2,\cdots,n^j_{l-t+1}]$ where $n^j_{i}$ denotes the number of tag $j$ in window $i$.
Next, for every sequence $S_{j}$, $j \in [1, k]$, the mean $\mu_j$ and the variance $\sigma_j$ of $S_{j}$ is calculated. We finally use every $\mu_j$ and $\sigma_j$ to construct the POS feature, namely the set of POS feature is $[\mu_1,\sigma_1,\mu_2,\sigma_2,\cdots,\mu_k,\sigma_k]$.


\subsection{Word features}
We also identify an interesting phenomenon on the titles of typical knowledgeable documents. Knowledgeable documents' title has a high probability using some conclusive words which indicates the documents will introduce concepts or properties of entities. Such as ``complete works of'' and ``encyclopedia'' indicate this document will introduce concept systematically, ``decrypt'' and ``guide'' indicate this document will introduce the relation of some entities from different aspects. Based on this observation, we concluded 25 keywords related to knowledge, and if a document's title contains any one of these keywords, the word feature will be set to $1$, otherwise $0$. Table~\ref{table:KnowledgeableWord} shows the detail of the words.


\begin{table}[htbp]\setlength{\tabcolsep}{5pt}
	\centering
	\caption{Conclusive words appear in titles.}\label{table:KnowledgeableWord}
	\begin{tabular}{c|c|c|c|c}
		\hline
		forum & real stuff & school & secret & skill \\
		\hline
		collection & question & decrypt & unscramble & guide \\
		\hline
		misunderstand & pattern & research & special & mechanism \\
		\hline
		classroom & difference & knowledge & comment & method \\
		\hline
		measure & answer & theory & study & system \\
		\hline
	\end{tabular}
\end{table}

\subsection{Sentence features}
The sentence features come from the statistics of shallow text. These statistics are usually used in traditional document classifications.
Because these features might also be predictive to knowledgeable and unknowledgeable documents, we put them into our feature set.
These features include the number of words, the length of document, the number of sentences and average length of sentences\cite{finn2006learning, tang2014text}. We also extend these features, such as the number of paragraphs, average sentences' number of each paragraph, the number of distinct words in title and so on. The results demonstrate that these features have more or less contributions on the accuracy of identifying knowledgeable documents. All the features in the three categories are shown as  Table~\ref{table:SentenceLevelFeature}.

\begin{table}[htbp]\setlength{\tabcolsep}{5pt}
	\centering
	\caption{Feature descriptions.}\label{table:SentenceLevelFeature}
	\begin{tabular}{l|l|l}
		\hline
		Features & No. & Description \\
		\hline
		POS & 1-98 & mean and variance of each POS \\
		\hline
		Word & 99 & number of conclusive words \\
		 & 100 & number of first personal pronoun \\
		 & 101 & number of second personal pronoun \\
		 & 102 & number of third personal pronoun \\
		 \hline
		Sentence & 103 & length of title \\
		 & 104 & length of content \\
		 & 105 & number of words in title \\
		 & 106 & number of words in content \\
		 & 107 & number of distinct words in title \\
		 & 108 & number of distinct words in content \\
		 & 109 & number of punctuation in title \\
		 & 110 & number of punctuation in content \\
		 & 111 & number of paragraphs in content \\
		 & 112 & number of sentences in content \\
		 & 113 & average number of sentences of paragraphs in content \\
		 & 114 & average number of words of sentences in content \\
		 \hline
	\end{tabular}
\end{table}

With the proposed features as shown in Table~\ref{table:SentenceLevelFeature}, we can train a binary support vector machine classifier~\cite{feldman2009part} to solute knowledgeable documents identification.



%% file: content/exp_zgb_xiang_v2.tex
\section{Experiments}\label{sec:experiment}

In this section, we verify the performance of the proposed models in a real-word dataset from Tencent Wechat public platform. First, we evaluate the effectiveness of the proposed models. Second, we demonstrate the advantages of SSNN on saving both time and memory consumption in training processes.

\subsection{Data Preparation}
We construct experimental datasets by the articles from Automobile, Finance and Real Estate domains in Wechat public platform.
For every domain, one thousand documents are manually labeled as knowledgeable or not according to the definitions in Section~\ref{sec:definition}, and we get three experimental datasets. Meanwhile, we build a mixture experimental dataset, denoted as AFR, which consists of the data from all the selected domains. Hence, we totally have four experimental datasets.
The details of the dataset are shown in Table \ref{table:datasetDescription}.
Then, every experimental dataset is divided into a training set and a test set, respectively. The training set contains 75\% of the samples, and the remains constitute the test set.


\begin{table}[htbp]
	\centering
	\caption{Datasets descriptions.}\label{table:datasetDescription}
	\begin{tabular}{c|c|p{0.66\columnwidth}}
		\hline
		\hline
		Name & KDR$^*$ & Content\\
		\hline
		Automobile & 19.0\% & News, knowledge of driving and cars.  \\
		\hline
		Finance & 12.9\% & Financial news, stock analysis and financial knowledge \\
		\hline
		Real Estate & 19.1\% &  News and advertising of real estate, knowledge of decoration  \\
		\hline
    AFR & 17.0\% & A mixture of Automobile, Finance and Real Estate. \\
    \hline
	\end{tabular}\\
	\tiny{
	$^{*}$``KDR'': Knowledgeable Documents Rate, the rate of knowledgeable documents in each domain.}
\end{table}

\subsection{Experimental Settings}

\subsubsection{Compared Methods}
There are all together twelve methods are compared in the experiments, and they can be categorized as follows. The notations and denotations of these methods are shown as Table~\ref{tb:methods}

\begin{itemize}
	\item \textbf{Convolutional neural networks methods.}
 Convolutional neural networks trained by individual experimental dataset are compared here. Both average and 1-max pooling are adopted for these CNNs. Hence, there are eight CNN-based methods in the comparison.

	\item \textbf{SVM based feature engineering methods.} Two SVM approaches are compared here. The main difference between them are the feature set used. The first one utilizes the feature set introduced in Section~\ref{sec:svm}. While the other one simply uses the TF-IDF vector based on the naive bag-of-word model.

	\item \textbf{The proposed SSNN methods.} Since SSNN can predict knowledgeability for different domains simultaneously, we just train one SSNN with the training set of AFR and perform predictions on every experimental dataset. Similar with the methods based on CNN, two pooling techniques are also conducted here. Hence, we get two versions of SSNN.

\end{itemize}

\begin{table}[htbp]\centering
	\caption{Compared Methods.}\label{tb:methods}
	\begin{tabular}{c|c|c|c}
		\hline
		\hline
		Method  & Training Set Domain & Pooling & Feature\\
		\hline
		CNN$_{\mathtt{avg}}$(Auto) &  Automobile & average & -  \\
		\hline
		CNN$_{\mathtt{avg}}$(Finance) & Finance & average & - \\
		\hline
   CNN$_{\mathtt{avg}}$(RE) & Real Estate & average & -  \\
		\hline
   CNN$_{\mathtt{avg}}$(AFR) & AFR & average & -  \\
		\hline
   CNN$_{\mathtt{max}}$(Auto) & Automobile & 1-max & -  \\
		\hline
CNN$_{\mathtt{max}}$(Finance) & Finance & 1-max & -  \\
		\hline
CNN$_{\mathtt{max}}$(RE) & Real Estate & 1-max & -  \\
		\hline
CNN$_{\mathtt{max}}$(AFR) & AFR & 1-max & -  \\
		\hline
SSNN$_{\mathtt{avg}}$ & AFR & average & -  \\
		\hline
SSNN$_{\mathtt{max}}$ & AFR & 1-max & -  \\
		\hline
SVM$_{\mathtt{TFIDF}}$ & AFR & - & TDIDF  \\
		\hline
SVM$_{\mathtt{FE}}$ & AFR & - & Feature set refer to Sec.~\ref{sec:svm}. \\
		\hline
	\end{tabular}
\end{table}


\subsubsection{Parameter Settings}\label{sec:imp}

For all compared neural networks, we build them with the help of Theano~\cite{bastien2012theano, bergstra2010theano}. Specifically, the mini-batch size is set to 10, the number of epoch is 10, the learning rate is set to 0.1.

The number of convolution kernel of CNN$_1$ is set to 50, the size of convolution kernel of CNN$_1$ is (200, 5). Hence, under this setting, we skip the sentences less than 5 words in CNN$_1$.
The number of convolution kernel of CNN$_2$ is set to 10, and the size of convolution kernel of CNN$_2$ is set to (10, 3). 
The number of nodes of fully connected layer is 10.

For the SVM$_{\mathtt{FE}}$, we empirically set the length of window $t=15$. Both SVM$_{\mathtt{FE}}$ and SVM$_{\mathtt{TFIDF}}$ are utilized linear kernels, and we set punishment coefficient $c=10$.

\subsection{Experimental Results}
Table~\ref{tab:expROC} and Table~\ref{tab:expAccuracy} demonstrate the performance of compared methods over ROC and prediction accuracies. Every compared model is tested on four different experimental datasets, namely Automobile, Finance, Real Estate and AFR.
\begin{table}[htbp]
	\centering
	\caption{The ROC of different methods.}\label{tab:expROC}
	\begin{tabular}{l|c|c|c|c}
		\hline
		\hline
		Data set & Automobile & Finance & RE & AFR \\
		\hline
		\rowcolor{mygray}
		CNN$_{\mathtt{avg}}$(Auto) & $0.9211$ & $0.8277$ & $0.8139$ & $0.8510$ \\
		CNN$_{\mathtt{avg}}$(Finance) & $0.8726$ & $0.9060$ & $0.6536$ & $0.8169$ \\
		\rowcolor{mygray}
		CNN$_{\mathtt{avg}}$(RE) & $0.9256$ & $0.7368$ & $0.8580$ & $0.8495$ \\
		CNN$_{\mathtt{avg}}$(AFR) & $0.9271$ & $0.8926$ & $0.8686$ & $0.9003$\\		
		\rowcolor{mygray}
		CNN$_{\mathtt{max}}$(Auto) & $0.9042$ & $0.7890$ & $0.7516$ & $0.8252$ \\
		CNN$_{\mathtt{max}}$(Finance) & $0.8636$ & $0.8700$ & $0.6931$ & $0.8152$ \\		
		\rowcolor{mygray}
		CNN$_{\mathtt{max}}$(RE) & $0.8836$ & $0.7113$ & $0.8392$ & $0.8230$ \\
		CNN$_{\mathtt{max}}$(AFR) & $0.9016$ & $0.8489$ & $0.8426$ & $0.8727$ \\
		\rowcolor{mygray}
		SSNN$_{\mathtt{avg}}$ & $\bm{0.9286}$ & $\bm{0.9087}$ & $\bm{0.8662}$ & $\bm{0.9015}$ \\
		SSNN$_{\mathtt{max}}$ & $0.8846$ & $0.8560$ & $0.8556$ & $0.8684$ \\
		\rowcolor{mygray}
		SVM$_{\mathtt{TFIDF}}$ & $0.7492$ & $0.6994$ & $0.7754$ & $0.8492$ \\
		SVM$_{\mathtt{FE}}$ & $0.9111$ & $0.8661$ & $0.8205$ & $0.8811$ \\
		\hline
	\end{tabular}
\end{table}

\begin{table}[htbp]
	\centering
	\caption{The accuracy of different methods.}\label{tab:expAccuracy}
	\begin{tabular}{l|c|c|c|c}
		\hline
		\hline
		Data set & Auto & Finance & RE & AFR \\
		\hline
		\rowcolor{mygray}
		CNN$_{\mathtt{avg}}$(Auto) & $0.8589$ & $0.8680$ & $0.8099$ & $0.8459$ \\
		CNN$_{\mathtt{avg}}$(Finance) & $0.8266$ & $0.8880$ & $0.6570$ & $0.7919$ \\
		\rowcolor{mygray}
		CNN$_{\mathtt{avg}}$(RE) & $0.8589$ & $0.8680$ & $0.8140$ & $0.8473$ \\
		CNN$_{\mathtt{avg}}$(AFR) & $0.8629$ & $0.8840$ & $0.8140$ & $0.8541$ \\		
		\rowcolor{mygray}
		CNN$_{\mathtt{max}}$(Auto) & $0.8619$ & $0.8683$ & $0.7953$ & $0.8428$ \\
		CNN$_{\mathtt{max}}$(Finance) & $0.8190$ & $0.8601$ & $0.7535$ & $0.8129$ \\		
		\rowcolor{mygray}
		CNN$_{\mathtt{max}}$(RE) & $0.7952$ & $0.8683$ & $0.8093$ & $0.8263$ \\
		CNN$_{\mathtt{max}}$(AFR) & $0.8524$ & $0.8642$ & $\bm{0.8419}$ & $0.8533$ \\
		\rowcolor{mygray}
		SSNN$_{\mathtt{avg}}$ & $\bm{0.8790}$ & $\bm{0.9000}$ & $0.8264$ & $\bm{0.8689}$ \\
		SSNN$_{\mathtt{max}}$ & $0.8629$ & $0.8800$ & $0.8306$ & $0.8581$ \\
		\rowcolor{mygray}
		SVM$_{\mathtt{TFIDF}}$ & $0.8336$ & $0.8349$ & $0.8295$ & $0.8403$\\
		SVM$_{\mathtt{FE}}$ & $0.8240$ & $0.8560$ & $0.8306$ & $0.8531$ \\
		\hline
	\end{tabular}
\end{table}

From these tables, we first can observe that the proposed SSNN surpasses other baselines on most of the cases. The only exception is CNN$_{\mathtt{max}}$(AFR) achieves the best performance on the Real Estate experimental set over the classification accuracy measure. On the other hand, the average pooling performs better than the 1-max pooling in SSNN on both two measures, we conjecture that the reason is average pooling keeps the semantic of context at a certain extent, which is beneficial to knowledgeable document identification.

Then, we look closer to the results of CNN based methods. They actually perform satisfactorily in their own training domains.
However, when they are applied to different domains, such as conducting CNN$_{\mathtt{avg}}$(Finance) in Real Estate, the performance slips significantly. This suggests that in experiments, the neural network trained by only one domain data are lack of generalization ability to another domain.
In addition, there are few differences between the ``avg'' models versus corresponding ``max'' models of CNN on the prediction accuracy measure. However, on ROC, except CNN$_{\mathtt{avg}}$(Finance)~(0.6536) versus CNN$_{\mathtt{max}}$(Finance)~(0.6931) in Real Estate, the ``avg'' models hold higher ROC over all the couples of models.

Next, we review the results of SVM-related models. Since we carefully summarize multiple features for the problem of knowledgeable documents extraction, SVM$_{\mathtt{FE}}$ outperforms SVM$_{\mathtt{TFIDF}}$ on most of the cases. However, compared with the neural network models, the approaches based on SVM are still weaker because of the reasons on feature selection or model generalization.

\begin{figure}
	\centering
	\subfigure[Roc curves.]{
		\includegraphics[width=3in]{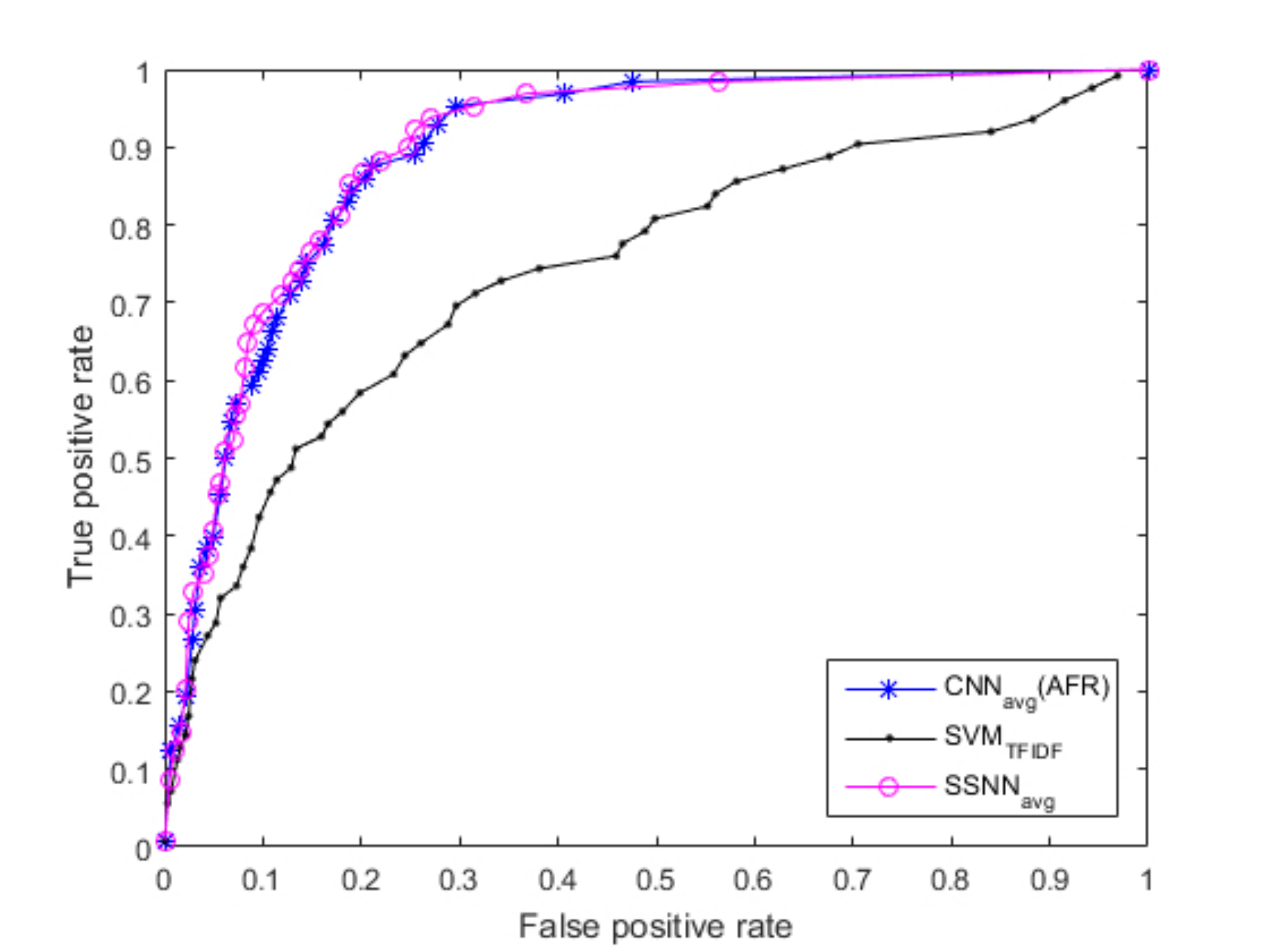}}
	\hspace{-0.3in}
	\subfigure[Zoom in top left corner.]{
		\includegraphics[width=3in]{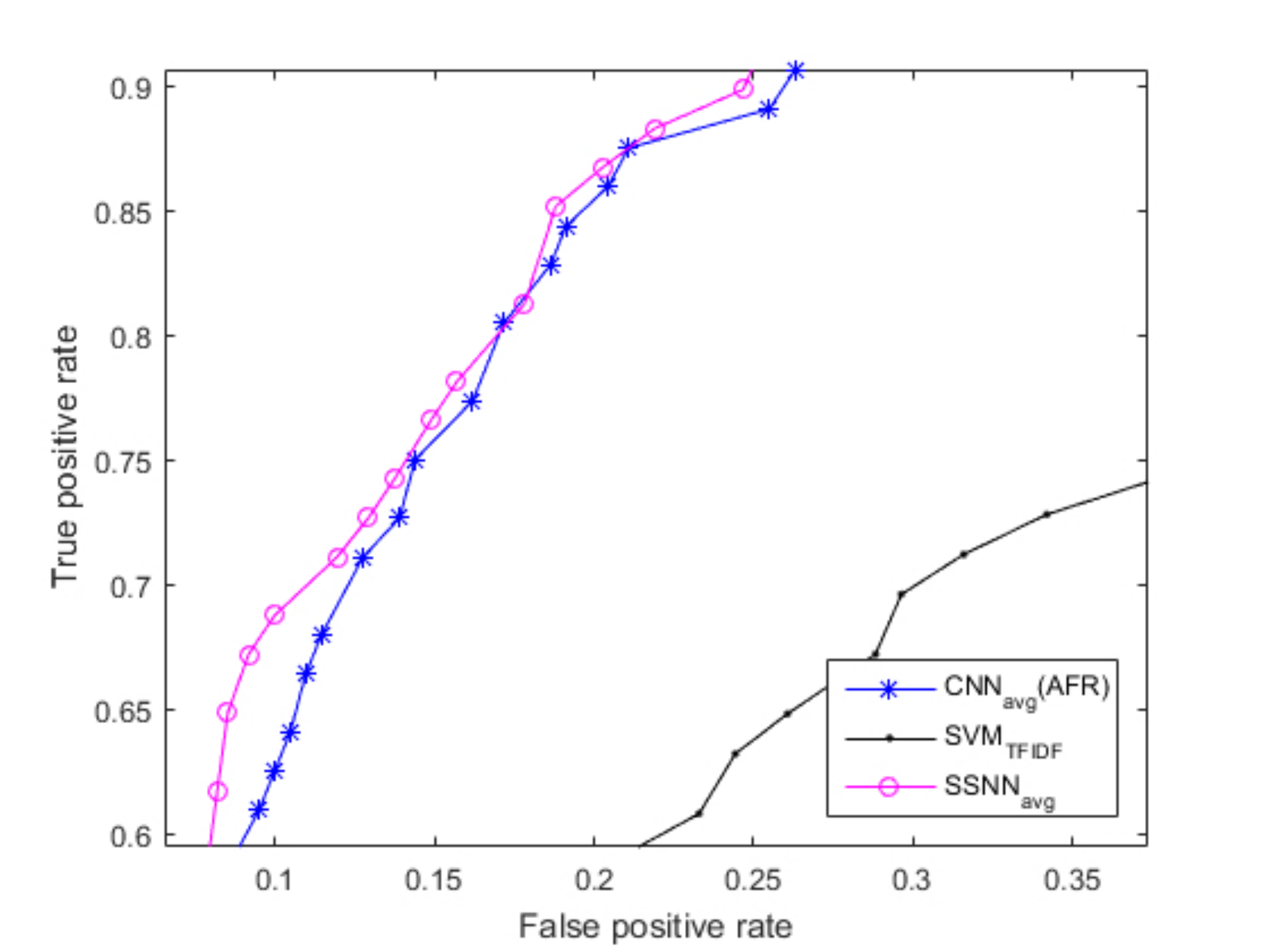}}
	\caption{Roc curves over the experimental set AFR.}
	\label{fig:ROC_curve} 
\end{figure}

Finally, to further demonstrate the performance of SNN-avg, the ROC curve in the experimental set AFR is shown as Fig~\ref{fig:ROC_curve}. From this figure, we observe
that the curve of SSNN$_\mathtt{avg}$ is leading the other models at most positions. It indicates that SSNN$_\mathtt{avg}$ performs better
than other models in such experimental set.

\subsubsection{Knowledgeable Snippets Extraction}
As we have discussed in section \ref{sec:extracSentences}, the Eq.~\ref{eq:formula} is able to extract knowledgeable sentences. Hence, we use such formula to conduct knowledgeable extraction in the experimental datasets. We discover some typical knowledgeable snippets and unknowledgeable ones as shown in Table~\ref{table:knowledgeableSentences}. Their saliency scores are highest in the documents where they appear. Observed from the table, the main characteristic of the knowledgeable sentences is that they define some concepts or discuss properties of entities and relations, while the unknowledgeable sentences mainly talk about advertisements or news which is not durable. The result is consistent with the definitions in Section~\ref{sec:definition}, which indicates that our model is also able to annotate the knowledgeable snippets.

\begin{table}[htbp]\setlength{\tabcolsep}{3pt}\small
	\centering
	\caption{Extraction results of knowledgeable or not-knowledgeable sentences.}\label{table:knowledgeableSentences}
	\begin{tabular}{c|p{0.4\columnwidth}|p{0.4\columnwidth}}
		\hline
		\hline
		Domain & Knowledgeable & Unknowledgeable Sentences\\
		\hline
		Automobile & When drive in the rainstorm, switching between low and high beam lamps is in favor of discovering the obstacles in the front. & There will be super unbelievable discounts in Chevrolet cars this summer, Biaoyu, Shenzhen city.  \\
		\hline
		Finance & Controling data means integrating the industrial chains. & Last week, 79 banks public launched 858 financial products. \\
		\hline
		Real Estate & Generally speaking, ceramic valve cores are durable, have high and low-temperature resistance, abrasion resistance and corrosion resistance. &  4000 hardbound rooms have been completed in 5 communities, Jiazhaoye, Shenzhen.   \\
		\hline
	\end{tabular}\\
\end{table}

We also excerpt an example of knowledgeable documents (translated from Chinese) as shown in Fig.\ref{fig:exp_sample2}. Sentences with the highest knowledgeable scores are colored blue, while sentences with the lowest scores red. A sentence is more knowledgeable if its score is higher.
The numbers in the squares denote the scores of the corresponding sentences.

This document demonstrates the concept of ``road rage''. The proposed model highlights five sentences in the excerpt:

\begin{figure}
	\centering	\fbox{\includegraphics[width=5in]{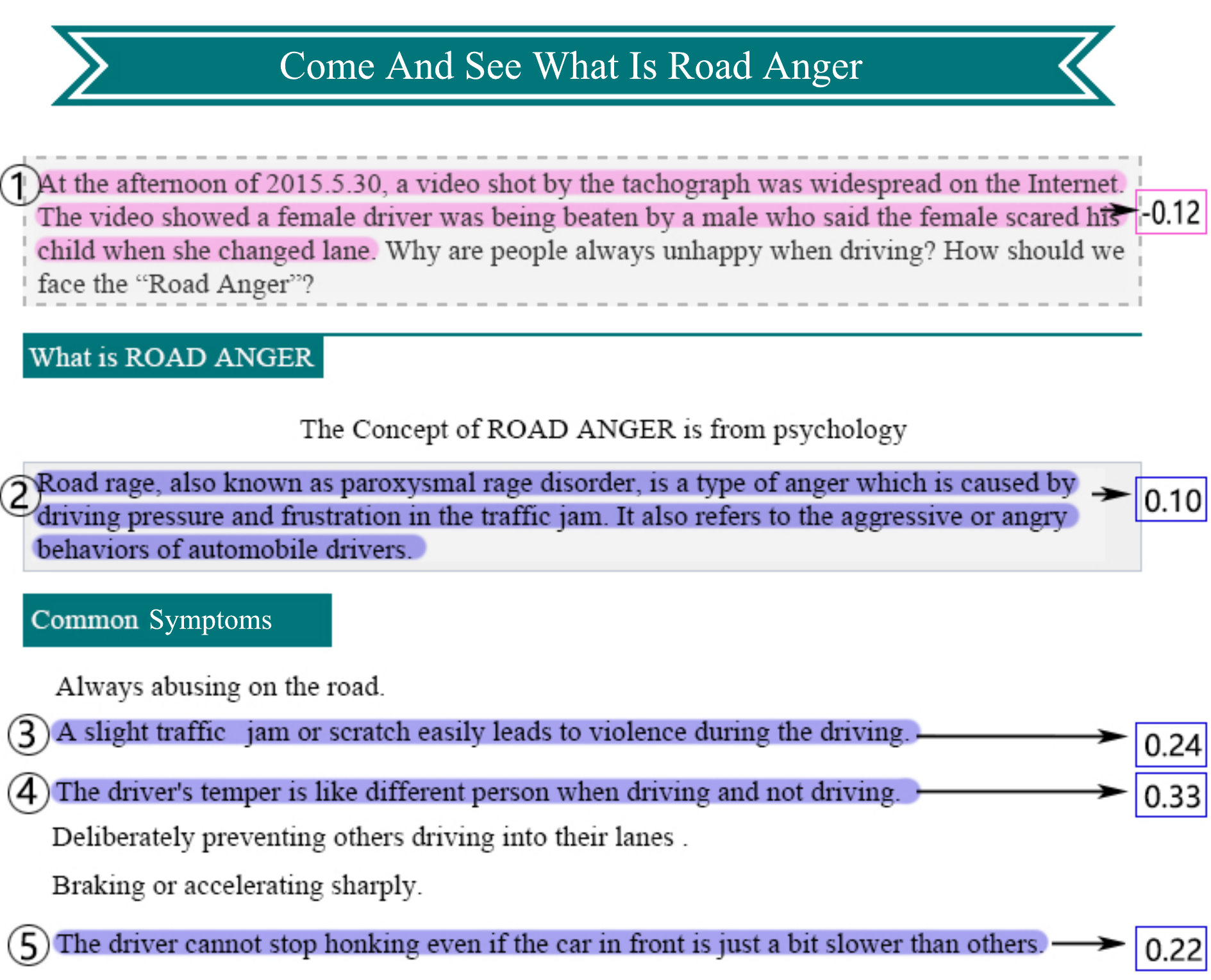}}	
	\caption{Example of knowledgeable snippet extraction --- Come and see what is road rage.}
	\label{fig:exp_sample2}
\end{figure}

\begin{itemize}
	\item[\ding{172}] ``\emph{At the afternoon of 2015.5.30, a video shot by the tachograph was wide-spread on the Internet. The video showed a female driver was being beaten by a male who said the female scared his child when she changed lane.}''  (score=-0.12)

This sentence is unknowledgeable, because it just describes the main content of an online video.
		
	\item[\ding{173}] ``\emph{Road rage, also known as paroxysmal rage disorder, is a type of anger which is caused by driving pressure and frustration in the traffic jam. It also refers to the aggressive or angry behaviors of automobile drivers.}''  (score=0.1)

This sentence is knowledgeable since it defines the ``\emph{road rage}''.
	
	\item[\ding{174}] ``\emph{The driver's emotion is usually uncontrolled, and a slight traffic jam or scratch easily leads to violence during the driving.}'' (score=0.24)

This sentence talks about two symptoms of whom have a road rage, namely ``the emotion is usually uncontrolled'' and ``traffic jam or scratch easily leads to violence''.
	
	\item[\ding{175}] ``\emph{The driver's temper is like different person when driving and not driving.}'' (score=0.33)

The sentence introduces another symptom of road range, which is knowledgeable.

	\item[\ding{176}] ``\emph{The driver cannot stop honking even if the car in front of him is just a bit slower than others.}'' (score=0.22).

Again, another symptom of road range is discussed in this sentence, which is knowledgeable as well.

\end{itemize}

Observed from those extracted knowledgeable sentences in experiments, we empirically believe they could conform to the definitions in Section \ref{sec:definition} satisfactorily.

%% file: conclusions/v1_zgb.tex
\section{Conclusions}\label{sec:conclusion}
In this paper, we introduce the concepts of knowledgeable snippets and documents, and formulate the problem of annotating knowledgeable documents and snippets. To solve the problem, we proposed a novel SSNN method and a feature engineering method to identify those knowledgeable documents from web data. In addition, SSNN can be further utilized to annotate knowledgeable snippets from documents.
The experiments on real data from Wechat public platform demonstrate the effectiveness of the proposed models.

\section{Acknowledgements}
The research work was supported by the National Key Research and Development Program of China under Grant No. 2017YFB1002104, the National Natural Science Foundation of China under Grant No. 91546122, 61573335, 61602438, 61473274, Guangdong provincial science and technology plan projects under Grant No. 2015 B010109005.




%% file: knowledgeable-snippet.bbl
\begin{thebibliography}{10}

\bibitem{auer2007dbpedia}
S{\"o}ren Auer, Christian Bizer, Georgi Kobilarov, Jens Lehmann, Richard
  Cyganiak, and Zachary Ives.
\newblock Dbpedia: A nucleus for a web of open data.
\newblock In {\em ISWC/ASWC}, pages 722--735, 2007.

\bibitem{baehrens2010explain}
David Baehrens, Timon Schroeter, Stefan Harmeling, Motoaki Kawanabe, Katja
  Hansen, and Klaus-Robert M{\"u}ller.
\newblock How to explain individual classification decisions.
\newblock {\em JMLR}, 2010.

\bibitem{bastien2012theano}
Fr{\'{e}}d{\'{e}}ric Bastien, Pascal Lamblin, Razvan Pascanu, James Bergstra,
  Ian~J. Goodfellow, Arnaud Bergeron, Nicolas Bouchard, and Yoshua Bengio.
\newblock Theano: new features and speed improvements.
\newblock In {\em NIPS Workshop}, 2012.

\bibitem{bergstra2010theano}
James Bergstra, Olivier Breuleux, Fr{\'e}d{\'e}ric Bastien, Pascal Lamblin,
  Razvan Pascanu, Guillaume Desjardins, Joseph Turian, David Warde-Farley, and
  Yoshua Bengio.
\newblock Theano: a cpu and gpu math expression compiler.
\newblock In {\em Proceedings of the Python for scientific computing conference
  (SciPy)}, 2010.

\bibitem{NIPS2011_4432}
Andrew Cotter, Ohad Shamir, Nathan Srebro, and Karthik Sridharan.
\newblock Better mini-batch algorithms via accelerated gradient methods.
\newblock In {\em NIPS}, pages 1647--1655, 2011.

\bibitem{Denil2015}
Misha Denil, Alban Demiraj, and Nando {De Freitas}.
\newblock Extraction of salient sentences from labelled documents.
\newblock In {\em ICLR}, 2015.

\bibitem{feldman2009part}
Sergey Feldman, Marius~A Marin, Mari Ostendorf, and Maya~R Gupta.
\newblock Part-of-speech histograms for genre classification of text.
\newblock In {\em IEEE ICASSP}, pages 4781--4784, 2009.

\bibitem{finn2006learning}
Aidan Finn and Nicholas Kushmerick.
\newblock Learning to classify documents according to genre.
\newblock {\em JASIST}, 2006.

\bibitem{Gatys2015}
Leon~A. Gatys, Alexander~S. Ecker, and Matthias Bethge.
\newblock A neural algorithm of artistic style.
\newblock {\em Nature Communications}, 2015.

\bibitem{Hearst1992}
M.~A. Hearst.
\newblock Automatic acquisition of hyponyms from large text corpora.
\newblock In {\em COLING}, pages 539--545, 1992.

\bibitem{Kalchbrenner2014}
Nal Kalchbrenner, Edward Grefenstette, and Phil Blunsom.
\newblock A convolutional neural network for modelling sentences.
\newblock In {\em ACL}, pages 655--665, 2014.

\bibitem{kessler1997automatic}
Brett Kessler, Geoffrey Numberg, and Hinrich Schütze.
\newblock Automatic detection of text genre.
\newblock In {\em ACL}, pages 32--38, 1997.

\bibitem{DBLP:journals/corr/Kim14f}
Yoon Kim.
\newblock Convolutional neural networks for sentence classification.
\newblock {\em CoRR}, 2014.

\bibitem{Krizhevsky}
Alex Krizhevsky, Ilya Sutskever, and Geoffrey~E Hinton.
\newblock Imagenet classification with deep convolutional neural networks.
\newblock In {\em NIPS}, pages 1097--1105, 2012.

\bibitem{Mikolov2013Distributed}
Tomas Mikolov, Ilya Sutskever, Kai Chen, Greg Corrado, and Jeffrey Dean.
\newblock Distributed representations of words and phrases and their
  compositionality.
\newblock In {\em NIPS}, pages 3111--3119, 2013.

\bibitem{suchanek2007yago}
Fabian~M Suchanek, Gjergji Kasneci, and Gerhard Weikum.
\newblock Yago: a core of semantic knowledge.
\newblock In {\em WWW}, pages 697--706, 2007.

\bibitem{DBLP:journals/corr/TaiSM15}
Kai~Sheng Tai, Richard Socher, and Christopher~D. Manning.
\newblock Improved semantic representations from tree-structured long
  short-term memory networks.
\newblock In {\em ACL}, pages 1556--1566, 2015.

\bibitem{tang2014text}
Peng Tang, Mingbo Zhao, and Tommy~WS Chow.
\newblock Text style analysis using trace ratio criterion patch alignment
  embedding.
\newblock {\em Neurocomputing}, 2014.

\bibitem{Tu2012Identifying}
Zhaopeng Tu, Yifan He, Jennifer Foster, Josef~Van Genabith, Qun Liu, and
  Shouxun Lin.
\newblock Identifying high-impact sub-structures for convolution kernels in
  document-level sentiment classification.
\newblock In {\em ACL}, pages 338--343, 2012.

\bibitem{Vinyals2014}
Oriol Vinyals, Alexander Toshev, Samy Bengio, and Dumitru Erhan.
\newblock Show and tell: A neural image caption generator.
\newblock In {\em CVPR}, 2015.

\bibitem{Wu2012Probase}
Wentao Wu, Hongsong Li, Haixun Wang, and Kenny~Q. Zhu.
\newblock Probase: a probabilistic taxonomy for text understanding.
\newblock In {\em ACM SIGMOD}, pages 481--492, 2012.

\bibitem{Xu2015}
K~Xu, J~Ba, and R~Kiros.
\newblock Show, attend and tell: Neural image caption generation with visual
  attention.
\newblock In {\em ICML}, pages 2048--2057, 2015.

\bibitem{zeiler2010deconvolutional}
Matthew~D Zeiler, Dilip Krishnan, Graham~W Taylor, and Rob Fergus.
\newblock Deconvolutional networks.
\newblock In {\em IEEE CVPR}, pages 2528--2535, 2010.

\bibitem{zeiler2011adaptive}
Matthew~D Zeiler, Graham~W Taylor, and Rob Fergus.
\newblock Adaptive deconvolutional networks for mid and high level feature
  learning.
\newblock In {\em IEEE ICCV}, pages 2018--2025, 2011.

\bibitem{zeki1993vision}
Semir Zeki.
\newblock A vision of the brain.
\newblock {\em Tex Dent J}, 1993.

\end{thebibliography}
